\documentclass{article}

\usepackage{PRIMEarxiv}

\usepackage[utf8]{inputenc} 
\usepackage[T1]{fontenc}    
\usepackage{hyperref}       
\usepackage{url}            
\usepackage{booktabs}       
\usepackage{amsfonts}       
\usepackage{nicefrac}       
\usepackage{lipsum}
\usepackage{fancyhdr}       
\usepackage{graphicx}       
\graphicspath{{media/}}     
\usepackage{float}  
\usepackage{setspace}  
\usepackage{titlesec} 
\usepackage{amsmath}

\titlespacing*{\section}{0pt}{10pt}{5pt}
\titlespacing*{\subsection}{0pt}{8pt}{4pt}
\titlespacing*{\subsubsection}{0pt}{6pt}{3pt}

\usepackage{setspace}
\title{Do Repetitions Matter? Strengthening Reliability in LLM Evaluations
\thanks{\textit{\underline{Citation}}: 
\textbf{Authors. Title. Pages.... DOI:000000/11111.}} 
}

\date{}

\usepackage{authblk} 
\usepackage{hyperref} 

\title{Do Repetitions Matter? Strengthening Reliability in LLM Evaluations
}

\author[1,2]{Miguel Angel Alvarado Gonzalez}
\author[1,2]{Michelle Bruno Hernandez}
\author[1,2,5]{Miguel Angel Peñaloza Perez}
\author[1,2,3]{Bruno Lopez Orozco}
\author[1,2,4]{Jesus Tadeo Cruz Soto}
\author[1,2]{Sandra Malagon}

\affil[1]{Carreras con Impacto}
\affil[2]{Aixo Lab}
\affil[3]{Facultad de Ciencias, UNAM, México}
\affil[4]{Facultad de Matemáticas, Universidad Veracruzana, México}
\affil[5]{Centro de Investigación Científica y de Educación Superior de Ensenada, Baja California, México}

\date{} 

\begin{document}
\maketitle

\begin{abstract}

LLM leaderboards often rely on single stochastic runs, but how many repetitions are required for reliable conclusions remains unclear. We re-evaluate eight state-of-the-art models on the AI4Math Benchmark with three independent runs per setting. Using mixed-effects logistic regression, domain-level marginal means, rank-instability analysis, and run-to-run reliability, we assessed the value of additional repetitions. Our findings shows that Single-run leaderboards are brittle: 10/12 slices (83\%) invert at least one pairwise rank relative to the three-run majority, despite a zero sign-flip rate for pairwise significance and moderate overall interclass correlation. Averaging runs yields modest SE shrinkage ($\sim$5\%
 from one to three) but large ranking gains; two runs remove $\sim$83\% of single-run inversions. We provide cost-aware guidance for practitioners: treat evaluation as an experiment, report uncertainty, and use $\geq 2$ repetitions under stochastic decoding. These practices improve robustness while remaining feasible for small teams and help align model comparisons with real-world reliability.

\end{abstract}


\section*{Introduction}

Large Language Model (LLM) evaluation helps ensure models operate as intended and builds trust among users and stakeholders. As LLM adoption in real-world applications accelerates (Cheng et al., 2025), the need grows for broader evaluations  that expose nuances in language understanding, generation quality, and task-specific proficiency (McGrath \& Jonker, 2024). Ultimately, integration of LLMs across scientific, industrial, commercial, and governmental settings depends not only on mean performance but also on the stability of outcomes under repeated trials.

LLM evaluations function as a quality-control layer: they guide technical advances while serving as an early-warning system for potential risks (Cao et al., 2025). It is therefore critical to design evaluations that report results in a methodologically sound manner—minimizing statistical noise and maximizing informativeness (Miller, 2024). Reliable evaluations do not merely indicate where LLMs can be deployed responsibly; they also shape how the community ranks the “best model” for a given task.

Despite the growth of benchmarking work, explicit experimental design around repetitions remains under-discussed. For instance, some groups (e.g., Epoch AI, 2025) run each model many times (often 8-16). Smaller research teams, however, may find such repetition counts infeasible. This creates a practical question with scientific consequences: how many repetitions are sufficient to obtain reliable conclusions without imposing prohibitive costs and operational challenges?

Prior suggestions to mitigate randomness include resampling answers, analyzing next-token probabilities, and computing question-level differences (Miller, 2024). Yet their applicability is often constrained by sample size. In our setting—an updated iteration of the AI4MAth Benchmark—we therefore complement these ideas with statistical tools that scale to modest budgets: sign-flip rate, intra-class correlation (ICC), and ANOVA, among others. We also expand the model set to include recent flagships—Grok-4, Claude Opus 4, and o3-mini—and, crucially, run each configuration three times to quantify run-to-run variability.

The aim of this work is to assess, within our benchmark and configurations, how many repetitions are necessary to achieve reliable results. We frame evaluation as an experiment rather than a leaderboard: we “evaluate our evaluations,” using statistical evidence to separate model signal from stochastic noise. In doing so, we draw on the long tradition of experimental design (Fisher, 1935; Box, Hunter \& Hunter, 2005; Montgomery, 2017) to motivate replication, variance estimation, and uncertainty reporting.
This focus is timely, as Arias et al. (2025) report a decline in the use of statistical validation in LLM evaluation papers, dropping from 32.6\% to 28.0\%. Moreover, when statistics are employed, the rigor is often insufficient Arias et al. (2025). Without a solid inferential basis, it becomes challenging to interpret task performance responsibly or to guide safe deployment. Repetition serves as a crucial tool: by quantifying how outcomes change across different runs, we can determine whether the observed differences are due to true capabilities or random chance.

Concerns about benchmark quality further underscore the need for robust methodology. Cao et al. (2025) synthesize recommendations—among them, repeated experiments and statistical validation—to reduce randomness and strengthen claims. In that spirit, we present a statistical exploration of whether repetitions materially affect conclusions in our benchmark and, if so, how many are needed for robustness. The resulting guidance informs future iterations of LLM evaluation under realistic resource constraints.

Finally, we view this work as part of the democratization of independent LLM evaluation. We consider this important because the development of diverse and open benchmarks enables a more representative and robust assessment of AI capabilities. Conversely, limited independent participation can create misaligned incentives and specification overfitting, where narrow metrics eclipse broader societal goals. Cost-aware replication strategies and accessible statistical tools are thus central to building evaluations that are both rigorous and widely adoptable.

\section*{Methodology}

This work is an iteration of the previous work of Peñaloza-Perez, et al (2025), who introduced the AI4Math benchmark dataset and settings that we reuse. The final dataset comprises 105 problems, balanced across the seven mathematical domains (see Table 1).

\begin{table}[H]
\centering
\begin{tabular}{lcc}
\hline
\textbf{Domain} & \textbf{Number of Problems} & \textbf{Percentage} \\
\hline
Algebra            & 15 & 14.29\% \\
Calculus           & 15 & 14.29\% \\
Combinatorics      & 14 & 13.33\% \\
Geometry           & 15 & 14.29\% \\
Mathematical Logic & 15 & 14.29\% \\
Number Theory      & 16 & 15.24\% \\
Probability        & 15 & 14.29\% \\
\hline
\textbf{Total}     & \textbf{105} & \textbf{100.00\%} \\
\hline
\end{tabular}

\vspace{6pt} 

\caption{Distribution of problems by mathematical domain. This table presents the distribution of the 105 problems in the AI4Math dataset across seven mathematical domains. Each problem includes a unique, exact answer and a detailed step-by-step solution. Percentages indicate the relative representation of each domain within the dataset.}
\label{tab:domain_distribution}
\end{table}

\subsection*{Evaluation Setup}

We evaluated eight state-of-the-art LLMs on the AI4Math benchmark under uniform experimental conditions. These models represent a mix of closed and open source systems, as well as general-purpose and reasoning-specialized architectures:  GPT-4o, GPT-4o mini, o3-mini, LLaMA 3.3 70B, DeepSeek-V3 685B, Claude Opus 4, Grok-3 and Grok-4.

\subsection*{Evaluation Conditions}

Language and Reasoning Setup: To assess the reasoning capabilities of large language models on AI4Math, we designed four evaluation conditions that vary along two axes: input language (Spanish vs. English) and direct answer (zero-shot, ZS) vs. Zero-Shot Chain-of-Thought (ZS-CoT). These conditions allowed us to isolate the effects of linguistic competence and step-by-step reasoning guidance. 

\subsection*{Answers Review}
All model outputs were manually reviewed by the authors. Final correctness was determined based solely on the match between the model’s final answer and the ground-truth solution.

\subsection*{Token Budget \& Other Settings}
All model responses were generated using a fixed decoding configuration across all conditions to ensure comparability. We used a temperature of 0.7 and $top-p$ of 0.95, selected to encourage coherent reasoning while allowing moderate variation in output style. In comparison to the first AI4Math Benchmark paper (Peñaloza-Pérez et al., 2025), we removed the maximum output length cap of 2,500 tokens for these iterations, as newer models such as Claude Opus 4, Grok-3, and Grok-4 weren’t able to produce an answer under this restriction.

\subsection*{Sampling}

Each model was evaluated in triplicate for every problem (three independent runs per model). This design was not only intended to mitigate randomness from stochastic decoding, but also to test whether repeated evaluations materially improve the robustness of performance estimates or whether single runs already provide stable conclusions.

\subsection*{Estimation \& Reporting}
Because the outcome is binary and measured repeatedly on the same problem instances across multiple runs, we used a \textbf{logistic mixed-effects model}. This allows subject/problem-specific random effects to capture unobserved heterogeneity (some problems are intrinsically easier/harder) and additional run-to-run variability for the same problem.

Models are estimated by maximum likelihood with a logit link
(\texttt{R} \texttt{lme4::glmer}, Laplace approximation, \texttt{bobyqa} optimizer).
We report odds ratios (OR) with 95\% Wald confidence intervals for fixed effects.
For model comparisons within language--prompt strata we use marginal means and Holm-adjusted
pairwise contrasts (\texttt{emmeans}). Descriptively, we report accuracy by
(model, language, prompt) with 95\% Wilson intervals.
 
Domain analysis added the factor domain (7 levels) to the same model; type‑III Wald $\chi^2$
 tests assessed main and interaction effects.

Run‑to‑run variability was measured as (a) flip‑rate—percentage of problems where a model’s answer changed across runs—and (b) intra‑class correlation (ICC) computed with irr::icc.

Rank‑instability compared single‑run leaderboards with the three‑run majority‑vote ground truth; a slice ($language * prompt$) was flagged unstable if any model pair inverted order.

Sign‑flip rate recorded whether pairwise significance decisions (CI excludes 0) changed when moving from one to three runs.

We quantified how the standard error (SE) of slice-level accuracy shrinks as R stochastic runs are averaged . For each model and slice ($language * prompt$), we computed per-run accuracy and the across-run SD; the predicted IID (Independent, Identically Distributed, IDD) curve was $SD\sqrt{R}$. The empirical SE was computed on a fixed slice set that contained all runs; for $R=2$, we averaged the SE over all run pairs ({1-2}, {1-3}, {2-3}), and for $R=3$, we used all three runs. Slice-level SEs were weighted by the mean items per run and then averaged to per-model and overall (model-averaged) curves, allowing us to assess near-IID behavior and quantify diminishing returns from additional runs.

The best model was selected through nested model comparison, using the likelihood ratio test (ANOVA) and the Bayesian Information Criterion (BIC).
All analyses were performed in R 4.3; code is released at \href{https://github.com/malvarado-tech/AI4MATH_2_R_project}{$https://github.com/malvarado-tech/AI4MATH\_2\_R\_project$}

\section*{Results \& Discussion}

This section reports the performance of eight large language models (LLM) on the AI4Math Benchmark across four evaluation conditions: Zero-Shot (ZS) and Zero-Shot Chain-of-Thought (ZS-CoT), each in English and Spanish.  It also features an accuracy comparison per domain and several statistical proofs to assess whether iterating repetitions on the benchmarks really provides any benefit to the robustness of the study.

\subsection*{Overall performance for each model and evaluation setting}

Overall performance was calculated using the success rate of the models across all evaluation settings. The benchmarks have a total of 105 questions, iterated in three repetitions, giving a total of 315 prompts in this experiment.

Across all 315 prompts per setting, Grok-4 achieved the highest aggregate accuracy (79.7\%, 95\% CI [74.9\%, 83.8\%]), followed by o3-mini (77.5\%, 95\% CI [72.5\%, 81.7\%]) and Claude Opus 4 (63.8\%, 95\% CI [58.4\%, 68.9\%]) (Figure 1).

\begin{figure}[H] 
    \centering
    \includegraphics[width=0.9\textwidth]{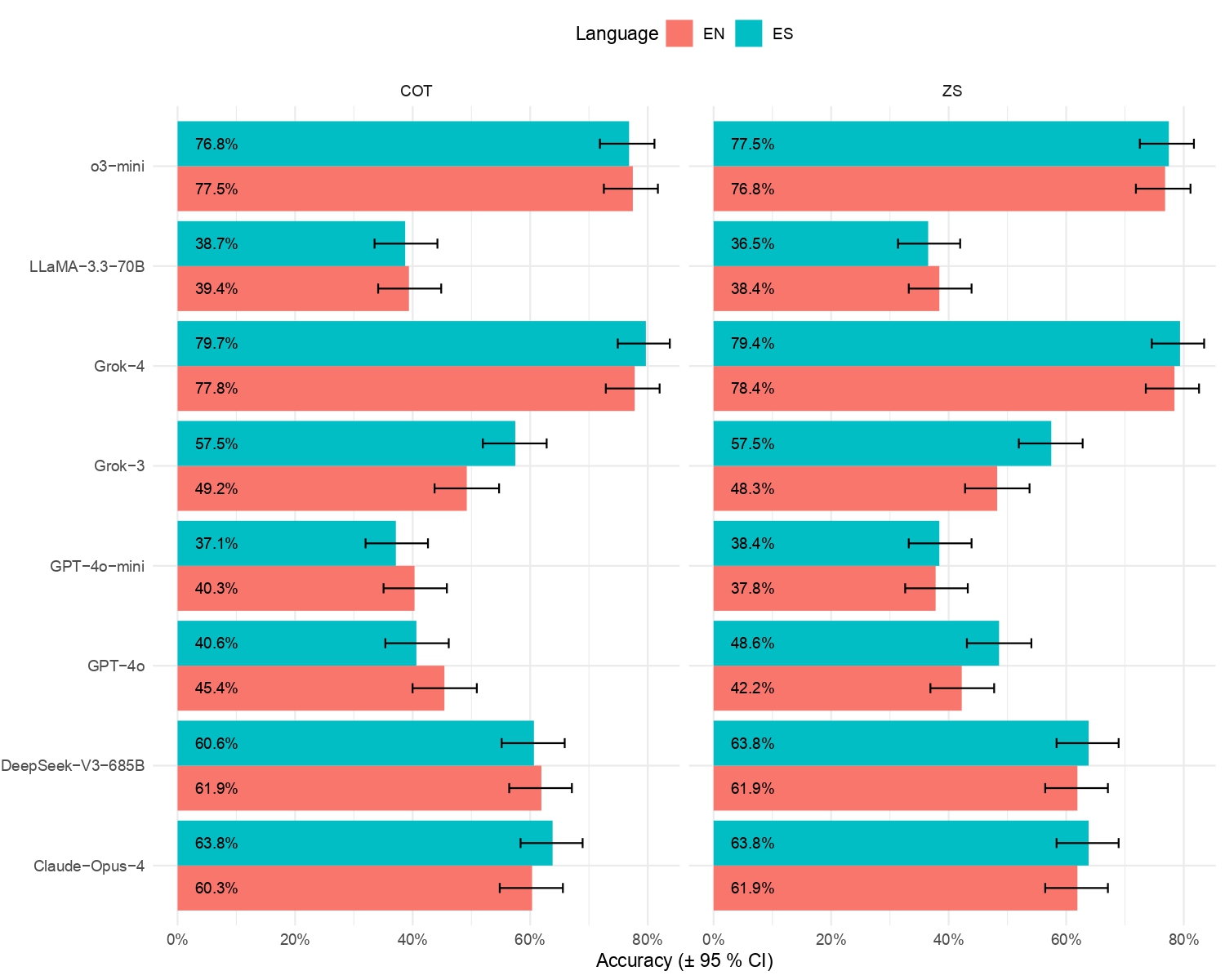} 
    \caption{Overall Performance per Model and Evaluation Setting. This figure presents the model names on the Y-axis, (o3-mini, Llama 3.3 70B, Grok 4, Grok 3, GPT-4o mini, GPT- 4o, DeepSeek V3 685B and Claude Opus 4) and the accuracy on the X-axis. For each model, it is categorized by Language (English in Soft-red, Spanish in Strong-Cyan) and by their evaluation setting in two columns (ZS-COT left, ZS Right). Overall accuracy was reported with Wilson 95\% CIs.}
    \label{fig:overall_performance}
\end{figure}

\subsection*{Mixed‑effects Logistic Regression}
\subsubsection*{Model Selection}

We compared five mixed‐effects logistic regression models varying the fixed‐effect structure (Figure 2). The model with the lowest BIC was model + language (BIC = 7691.4), which outperformed the next best model (model + language + prompt) with a $\Delta BIC$ of 9.0 and a BIC weight of 0.99. According to conventional guidelines ($\Delta BIC > 6$; Kass \& Raftery, 1995), this provides strong evidence in favor of the more parsimonious model + language specification. All models included random intercepts for $problem\_id$ and $problem\_id:run$. This result indicates that prompt did not contribute meaningfully to explaining variation in correctness beyond what was already captured by model and language, and therefore can be excluded from the final model.
\begin{figure}[H] 
    \centering
    \includegraphics[width=0.7\textwidth]{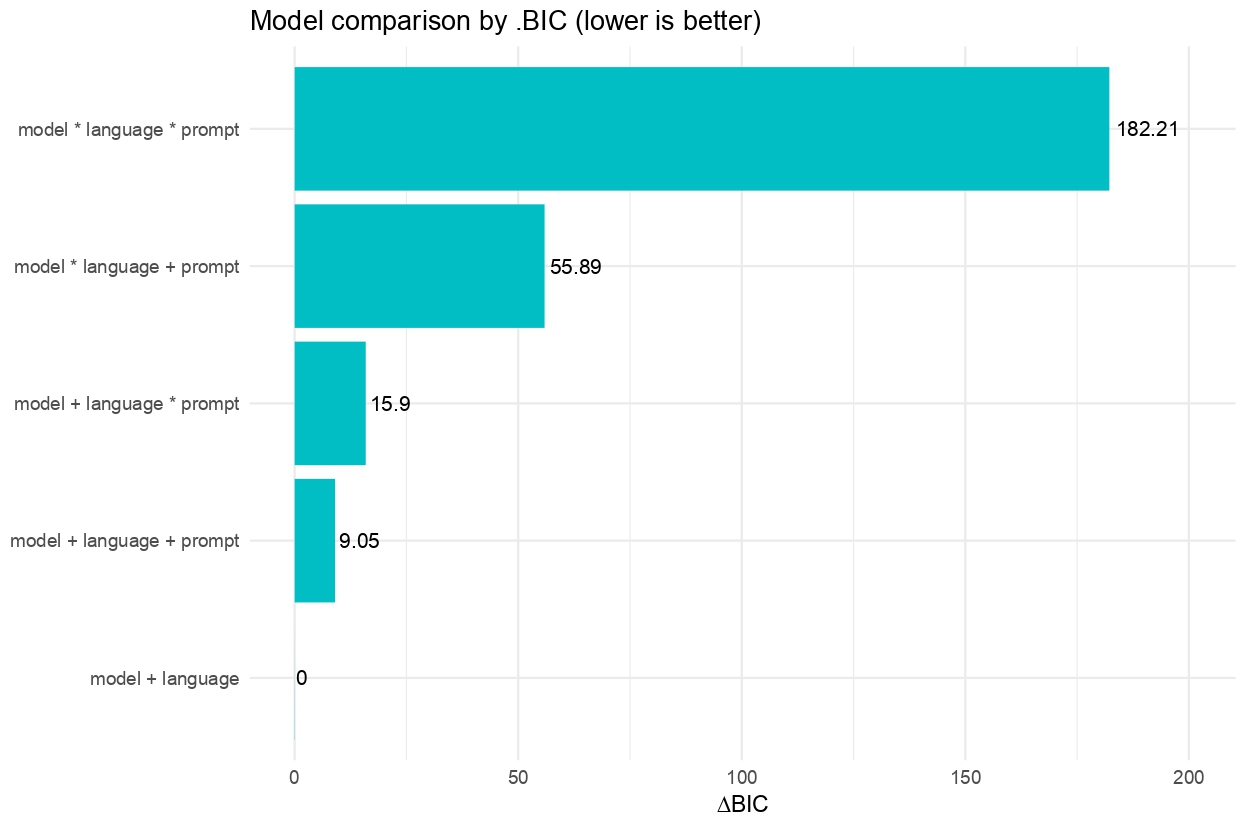} 
    \caption{ Model selection by $\Delta BIC$. The y-axis lists the five candidate mixed-effects logistic specifications ($model + language$, $model + language + prompt$, $model + language * prompt$, $model * language + prompt$, $model * language * prompt$). The x-axis shows $\Delta BIC$ relative to the best model (lower is better). The shortest bar ($\Delta BIC = 0$) corresponds to $model + language$; the next best is $model + language + prompt$ ($\Delta BIC = 9.0$; BIC weight = 0.99). Larger $\Delta BIC$ values for specifications that include prompt interactions indicate strong to very strong evidence against adding prompt terms. All models include random intercepts for $problem\_id$ and $problem\_id:run$. }
    \label{fig:model_selection}
\end{figure}
Figure 3 reports the estimated logs ratio for each language model in comparison to Grok 4, along with their 95\% confidence intervals. Values closer to 1 indicate higher estimated proportions of correct responses comparated to the reference, closer to 0 indicate lower proportions.

\begin{figure}[H] 
    \centering
    \includegraphics[width=0.7\textwidth]{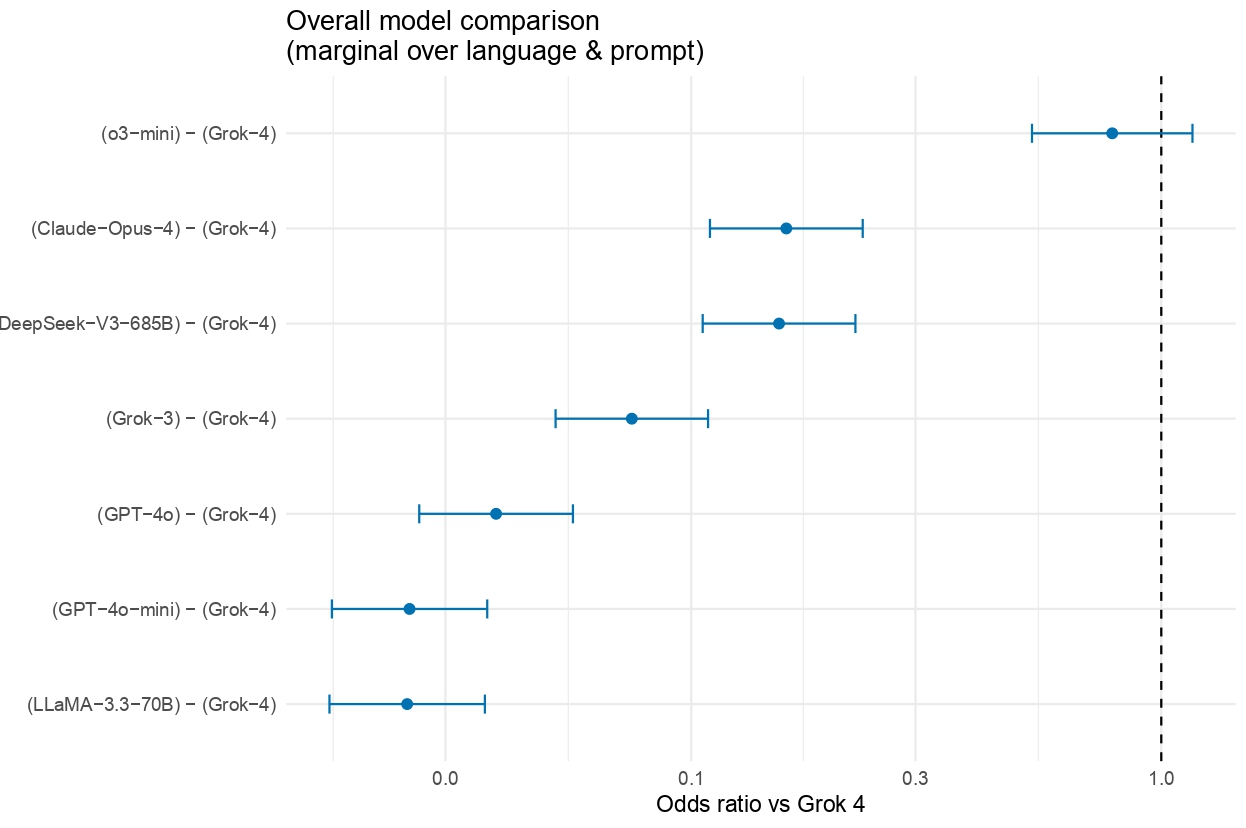} 
    \caption{Odds-Ratio from Mixed-effect logistics regression. This figure represents the overall model comparison contrasted with Grok 4. The x-axis represents the odds ratio expressed in log-10, the y-axis represents the model's pair of models under comparison.}
    \label{fig:odds_ratio}
\end{figure}

Mixed-effects logistic regression aggregated over language and prompt variants shows Grok-4 with the highest point estimate. The only challenger whose performance is statistically indistinguishable from Grok-4 is o3-mini  (OR=0.79, 95\%CI (0.53–1.17, $p=0.43$).All other models—including Claude-Opus-4, DeepSeek-V3-685B, and LLaMA-3-70B score lower on average.

Grok-4, o3-mini, and Claude-Opus-4 are the top three models by overall accuracy in this benchmark. To keep the conclusions durable as systems evolve, we emphasize effect sizes and uncertainty rather than specific version labels: among the leaders, gaps are modest and may shift as models change. These patterns are consistent across languages and prompts in this iteration.

We therefore suggest that future benchmarks move beyond raw accuracy and incorporate qualitative analyses of reasoning processes—how models arrive at their solutions, consistency across runs, and common error types—so evaluations capture not only whether a model is correct, but also how reliable and transparent the reasoning is.

\subsection*{Domain-Specific Analysis}
\subsubsection*{Model Selection - Domain Analysis}

We compared two nested mixed-effects logistic regression models varying the treatment of the domain (Supplementary Figure 1). The model with the lowest BIC was the additive specification, $model + domain (BIC = 7729.3)$, which outperformed the interaction alternative, $model \times domain$, with $\Delta BIC = 163.3$ and a BIC weight of 1.00. According to conventional guidelines $(\Delta BIC > 6)$, this provides very strong evidence in favor of the additive specification. All models included random intercepts for $problem\_id$ and $problem\_id:run$. This result indicates that domain does not meaningfully moderate model performance beyond a shared difficulty shift; the relative ordering of models is stable across domains.

The mixed‑effects logistic regression revealed significant main effects of both model $(\chi^2= 1217.0, df = 7, p < 2e-16)$ and  domain $(\chi^2 = 13.5, df = 6, p = 0.005)$. This justifies a domain‑level breakdown (Fig. 5) and shows that aggregate leaderboards obscure specialised strengths and weaknesses (including random intercepts for $problem\_id$ and $problem\_id:run$).

\subsubsection*{Accuracy per Domain} 

In the previous iteration of this work, Peñaloza-Perez et al (2025) described Geometry, Combinatorics and Probability as the most challenging domains for all models. To  assess whether these conditions changed in this examination, we calculated the marginal means for all domains. As mentioned, the Domain factor shares baseline difficulty equally across models, meaning the baseline difficulty of each domain is persistent in all the models tested in this study. Furthermore, baseline domain difficulty (marginal means) varies substantially and matches the findings previously reported by Peñaloza-Perez et al, (2025) being Geometry the hardest domain ($p= 29\%$) and Algebra the easiest ($p= 89\%$) (see figure 4).

\begin{figure}[H] 
    \centering
    \includegraphics[width=0.7\textwidth]{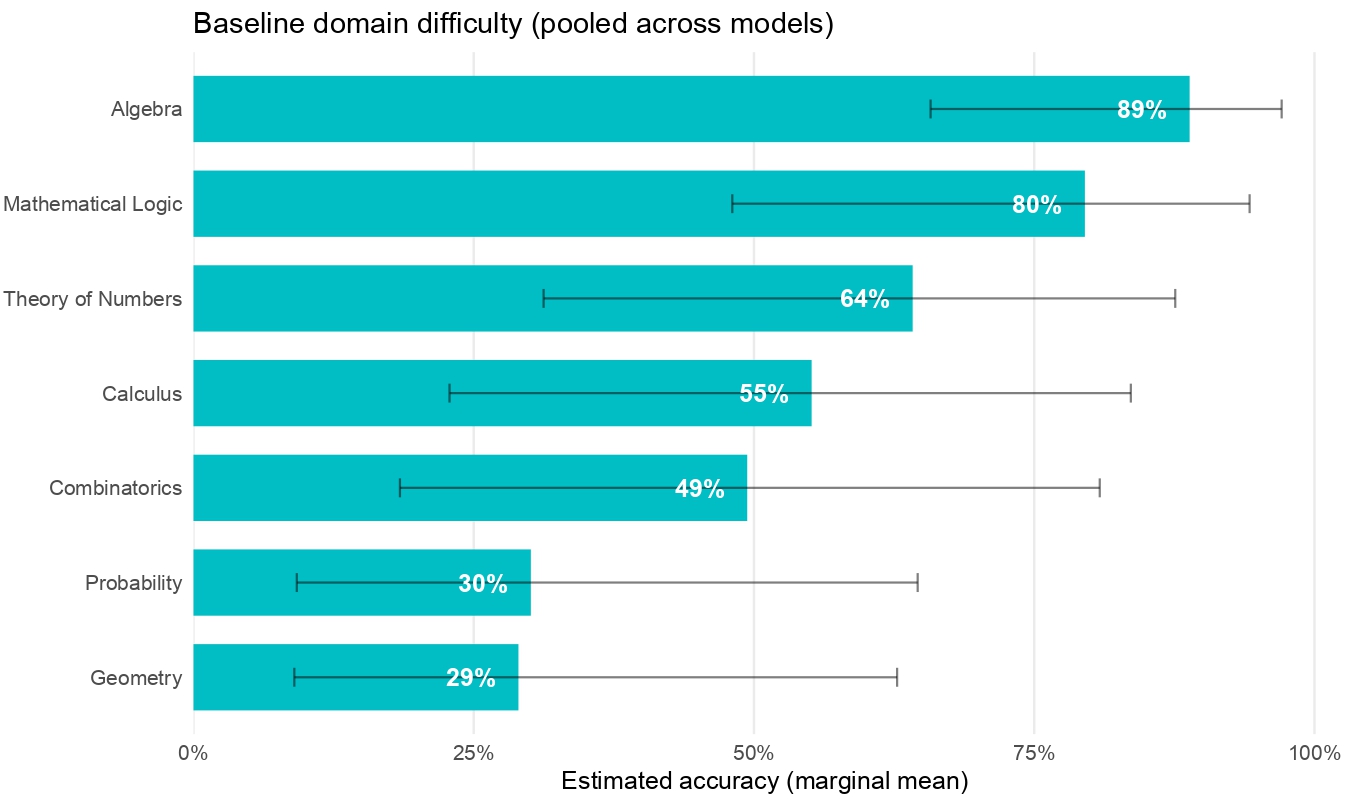} 
    \caption{Baseline domain difficulty (pooled across models). The y-axis lists mathematical domains; the x-axis shows the estimated accuracy (marginal mean, \%) from the additive mixed-effects logistic model (correct $\sim$ model + domain, random intercepts for $problem\_id$ and $problem\_id:run$). Bars give the pooled probability of a correct answer; whiskers are 95\% Wald CIs. Higher values indicate easier domains. Algebra ($\sim$89\%) and Mathematical Logic ($\sim$80\%) are easiest; Probability ($\sim$30\%) and Geometry ($\sim$29\%) are hardest, with Theory of Numbers ($\sim$64\%), Calculus ($\sim$55\%), and Combinatorics ($\sim$49\%) in between.
}
    \label{fig:domain_difficulty}
\end{figure}

Consistent with Figure 4, Figure 5 displays descriptive accuracies for each model within every domain; it illustrates absolute performance levels and the overall gradient of domain difficulty (not inferential tests).

\begin{figure}[H] 
    \centering
    \includegraphics[width=0.7\textwidth]{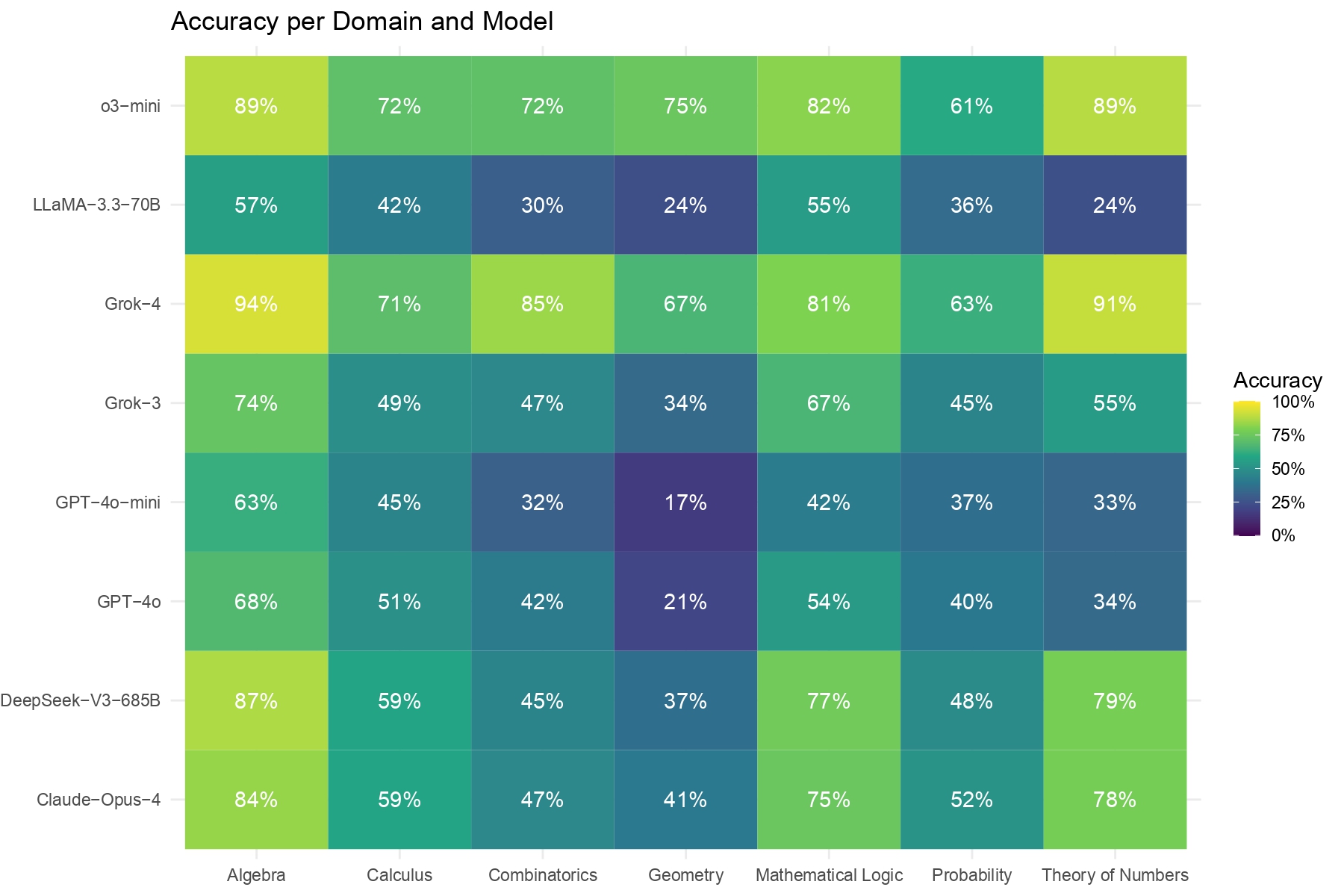} 
    \caption{Accuracy per Domain Heatmap. This heatmap represents the accuracy in percentage (\%) per domain and model across three stochastic runs. The x-axis enlist Math Domain, while the y-axis enlist the models. The intersection model/domain is coloured based on their performance, where cells are colored from lower accuracy (blue) to higher accuracy (yellow).
}
    \label{fig:domain_accuracy}
\end{figure}

Averaging across domains, Grok-4 attains the highest mean accuracy with comparatively low variation (78.7\%, SD = 12.0 pp; IQR = 18.9 pp; range 63.3–93.9\%), closely followed by o3-mini (77.0\%, SD = 10.2 pp; IQR = 13.5 pp; range 60.6–88.9\%) (Supplementary Table 7 Model Robustness Across Domains). This combination of high average accuracy and small cross-domain dispersion indicates robust, “generalist” performance. In contrast, the remaining models average $\leq$ 62\% and exhibit larger variability, suggesting greater domain sensitivity.

\subsection*{Qualitative Comparison of Grok 4 vs o3-mini}
Using the additive model (\textit{model + domain}), we estimated domain-conditional marginal means and contrasted each model against Grok-4 on the log-odds scale (Holm-adjusted \textit{p}-values). Converting contrasts to odds ratios (OR), every competitor shows $OR < 1$ across all seven domains—i.e., lower odds of a correct answer than Grok-4—and, after multiplicity control, almost all differences are statistically significant (blue markers; Supplementary Figure 2). Because the selected model excludes a $model * domain$ interaction, these contrasts are invariant across domains on the link scale. Among all competitors, o3-mini was not significantly lower than Grok-4; accordingly, we present a qualitative comparison between Grok-4 and o3-mini across our benchmark, highlighting patterns in solution strategies, failure modes, and problem types where one model holds an advantage. These two models were selected because, as noted above, they consistently led overall performance among the evaluated systems and are the only pair with relatively comparable performance, making them particularly suitable for a qualitative exploration of potential differences in reasoning approaches.

We’d identified a similar behavior as the one described by Burnham (2025), when Grok-4 “grinds out” the problem. For example, in a problem that asks to factor a polynomial, the solution can be intuited, the proposed roots tested, and verified. This problem has been solved consistently by all (or almost all) models. Grok-4 does not take risks and repeatedly applies synthetic division to obtain the decomposition. This coincides with a pattern of relying on general formulas and common procedures even when following these implies a lot of calculations. This behavior when compared to o3-mini could have given Grok-4 an advantage on some problems.
At the same time, there were thirteen problems that Grok-4 and o3-mini failed to solve correctly over all evaluation settings. In half of them, there was a specific step where the solution required to calculate areas, whether it is because parameters must be found that maximize the area of a geometric figure or because the problem consists of finding the area of a figure (or figures) given a limited amount of information. This suggests that state-of-the-art models might present weaknesses in calculating areas, or when constrained to low-information settings, or in certain optimization problems.

In contrast, o3-mini outperformed Grok-4 on eight mathematical problems related to spatial visualization. The ability to visualize geometric constructions helps in identifying relevant objects, cases, and premises necessary for formulating arguments and performing calculations to arrive at the correct answer. Exploring these capabilities further is worthwhile, as the specifications may be important depending on the different challenges or tasks we aim to undertake in the future. 

Finally, there were two problems that only Grok-4 was able to obtain the right answer in any of its independent runs (above all LLM’s tested during this study). The first one belongs to the Combinatorics domain, where understanding how to formulate a partition function suggests a key step into getting to the right answer. The second problem corresponds to the Mathematical Logic domain, where the ability to solve systems of linear equations lead to the correct answer.

Overall, these observations indicate complementary strengths that can depend on the problem’s demands—procedural persistence and formulaic grind for Grok-4, versus spatial visualization advantages for o3-mini—while also revealing shared weaknesses in area-related and low-information optimization tasks.

\subsection*{Run to Run Instability Analysis}

To assess run-to-run instability, we measured variability in two complementary ways: (a) flip-rate, defined as the percentage of problems where a model’s answer changed across runs, and (b) intra-class correlation (ICC) computed with $irr::icc$. Flip-rate captures the entropy of responses—indicating how often users would encounter different answers to the same question—and can be interpreted as a measure of “noise” at the question level. In contrast, ICC provides a more global reliability estimate, quantifying the consistency of model outputs across repeated trials.

\subsection*{Flip Rate (Entropy)}

\begin{figure}[H] 
    \centering
    \includegraphics[width=0.9\textwidth]{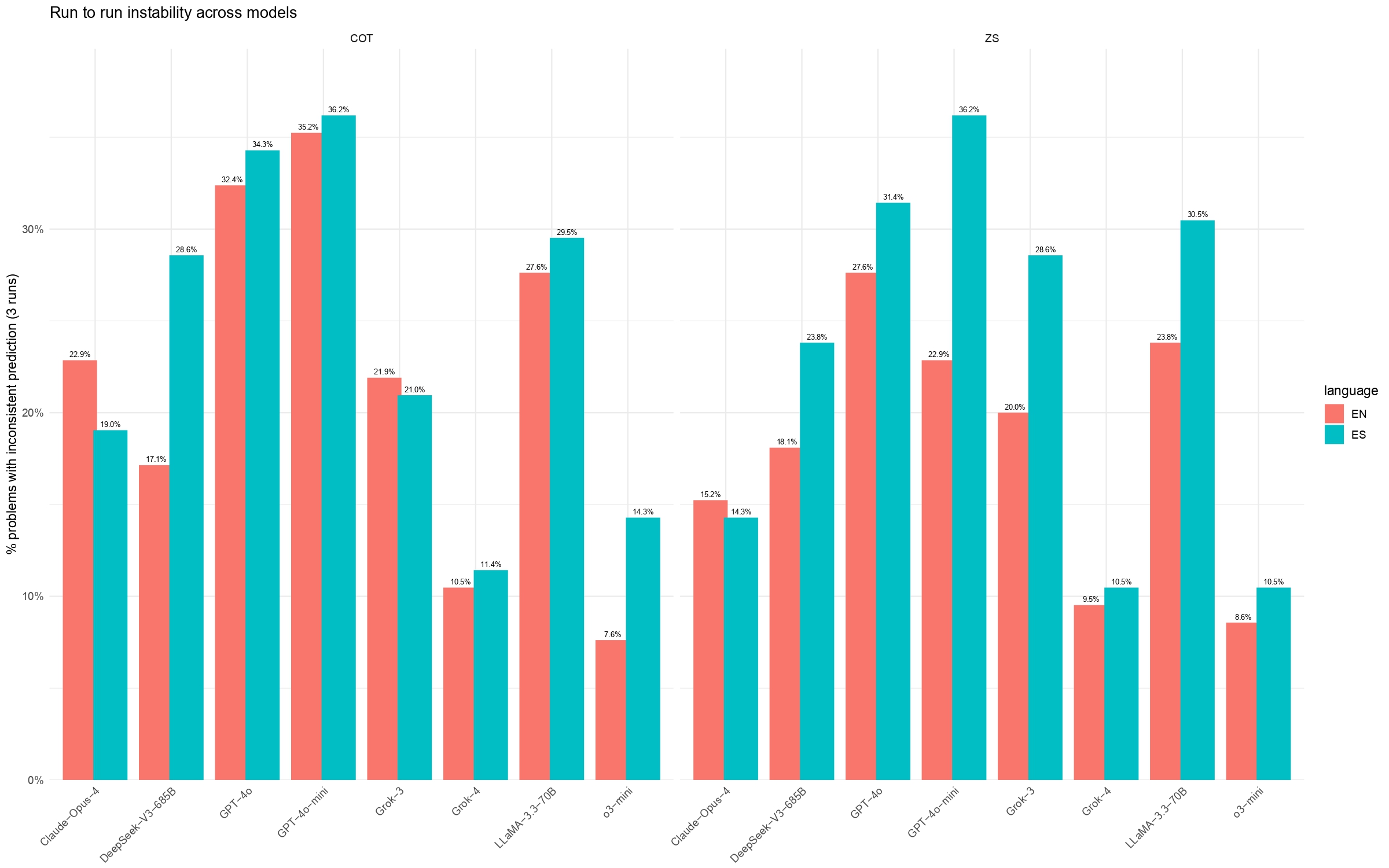} 
    \caption{Run-to-run instability across models. This figure shows model names on the X-axis and the percentage of problems with inconsistent prediction on the Y-axis (based on 3 repeated runs per problem).  Bars are grouped by language—English (soft red) and Spanish (strong cyan)—and split into two panels for the prompting condition (COT on the left, ZS on the right). Higher bars mean less run-to-run stability; these percentages summarize stability, not overall accuracy.
}
    \label{fig:run_instability}
\end{figure}

Because we decode with temperature 0.7 and top‑p 0.95, stochasticity is intentionally high. Under these settings GPT‑4 family models change their answer on $\sim$35\% of problems, whereas Grok‑4 does so on $\sim$10\%. Therefore we retain the three‑run protocol to average out sampling variance. For deterministic evaluations (e.g., production inference) a temperature closer to 0 or greedy decoding would align Grok‑4’s leaderboard lead with much lower run‑to‑run noise. In other words, if we rerun the benchmark under a more deterministic decoding setting, it would be expected for the models to increase their stability by reducing run-to-run noise. However, the relative ranking would be unchanged and further examination of this claim is still needed.
It has been discussed that selecting an adequate method for a specific task should be taken into account to tweak the performance, robustness and speed of the decoding (Shi, et al 2024). Perhaps, exploring additional decoding settings to increase the self-consistency might boost the performance of the models, as this is a behavior reported by Wang et al (2022) when benchmarking Chain of Thought  prompting.

\subsection*{Intraclass Correlation Coefficient (ICC)}
\begin{figure}[H] 
    \centering
    \includegraphics[width=0.7\textwidth]{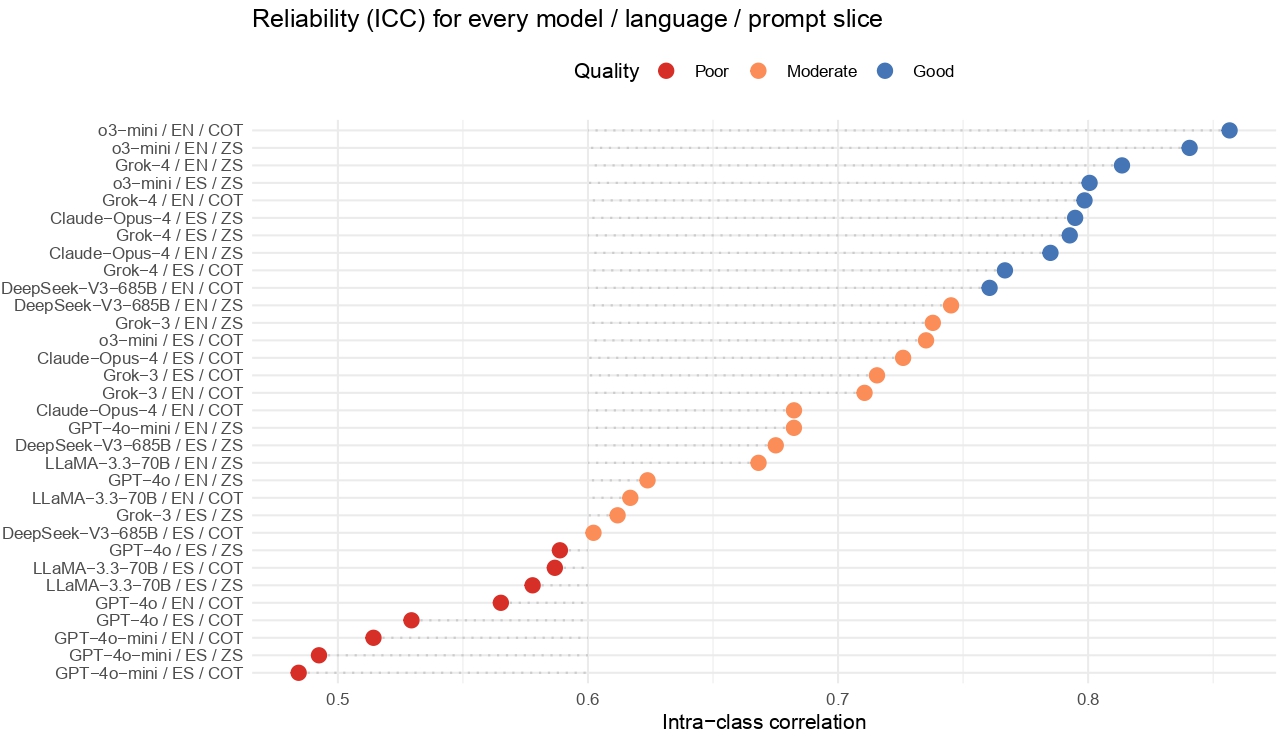} 
    \caption{Run-to-run Reliability (ICC) by Model, Language, and Prompt. This figure ranks model / language / prompt slices on the Y-axis and plots their Intra-class Correlation (ICC) on the X-axis, where larger values indicate more consistent outcomes across repeated runs on the same problems (1 = perfect agreement; 0 = chance). Each point is colored by a qualitative reliability label derived from the ICC: Poor (< 0.60, red), Moderate (0.60–0.75, orange), Good (0.75–0.90, blue); faint dashed guides mark the 0.60 threshold. ICCs are computed within each slice using a two-way, consistency, single-measure model (irr::icc). The plot summarizes reliability, not accuracy—higher ICC means the model’s correctness is more stable from run to run for that slice.
}
    \label{fig:icc}
\end{figure}

Slice‑wise ICC values (Figure 8) span 0.48–0.86. Grok‑4 and o3‑mini achieve good reliability in most of their settings ($ICC \geq 0.77$), indicating low run‑to‑run noise. By contrast, GPT‑4o‑mini falls into the poor band for both languages and prompting styles ($ICC \leq 0.52$), flipping answers on roughly half the problems. Overall, 10 of  32 slices rate as good, 14 as moderate, and 8 as poor, justifying our choice of three decoding repetitions at temperature of 0.7.

This analysis has also revealed interesting insights about how newer models have increased their self-consistency. Self-consistency and self-correction has been a going debate over recent years (Lie, et al 2024; Wang et al 2022). In this benchmark, Grok-4, o3-mini and Claude Opus 4 achieved a “good” consistency across the three samplings, while older models such as GPT-4 and LlaMa-3-70B exhibit unstable behavior on their sampling. There is evidence that factors such as zero temperature and fair prompting are very relevant to the “intrinsic ability” of LLMs to perform self-correction in their internal chain of thoughts (Liu et al, 2024). However, it is advised to perform thorough examination on the best decoding method for each specific use-case (Shi et al, 2024).

When evaluating LLM’s, it is important to produce results with high reliability so they serve as an accurate approximation of their real capabilities. In other areas, sampling methods are a standard for high quality research. Nonetheless, when doing research around LLM’s it is computationally expensive to sample every scenario to assure consistency in every evaluation setting. To approach this, some methodologies have been explored to reduce the need of sampling and increase the performance of LLM’s when performing specific tasks. For example, Taubenfeld et al (2025) propose the methodology of Confidence-Informed Self-Consistency (CISC) where  high-confidence reasoning paths are prioritized, matching the performance with only eight samples when compared with a thirty-sample self-consistency scenario. A similar approach was taken by Wan et al (2024) with Reasoning-Aware Self-Consistency (RASC) where reasoning paths are sampled and the most faithful rationale among generated samples is prioritized. Both approaches exhibit a systematic methodology to increase the performance of LLM's on specific tasks and feature promising results with expected reduced costs.

\subsection*{Rank Instability}

We quantify rank instability as the fraction of $language * prompt$ slices in which any pair of models swaps order relative to the three-run aggregate (“ground truth”); with 12 slices total, the instability rate is $p_{\text{inst}} = \text{\# slices with } \geq 1 \text{ inversion} / 12$. Using a single stochastic run, 10 of 12 slices (83\%) showed at least one rank inversion relative to the three-run aggregate. 

Thus, single-run leaderboards are fragile under our decoding settings; averaging three runs removes these inversions and yields stable rankings. Figure 8 visualizes the effect: line crossings indicate mismatches with the three-run ranking, whereas horizontal segments indicate agreement. (Ground truth is computed from three seeds; ties are resolved by mean accuracy. Decoding: temperature = 0.7, top-p = 0.95.). 

Therefore, this visualization highlights that in our benchmark a single run is insufficient to produce a stable or reliable ranking of LLMs. Averaging across three runs proved necessary to reach stability and generate a robust ranking. This provides evidence that repetitions are essential when constructing leaderboards, as such rankings often function as “quality-control” layers to guide and implement technical progress. Ensuring stability through multiple runs is thus critical for making informed decisions, rather than relying on results that may largely reflect randomness.

\begin{figure}[H] 
    \centering
    \includegraphics[width=0.8\textwidth]{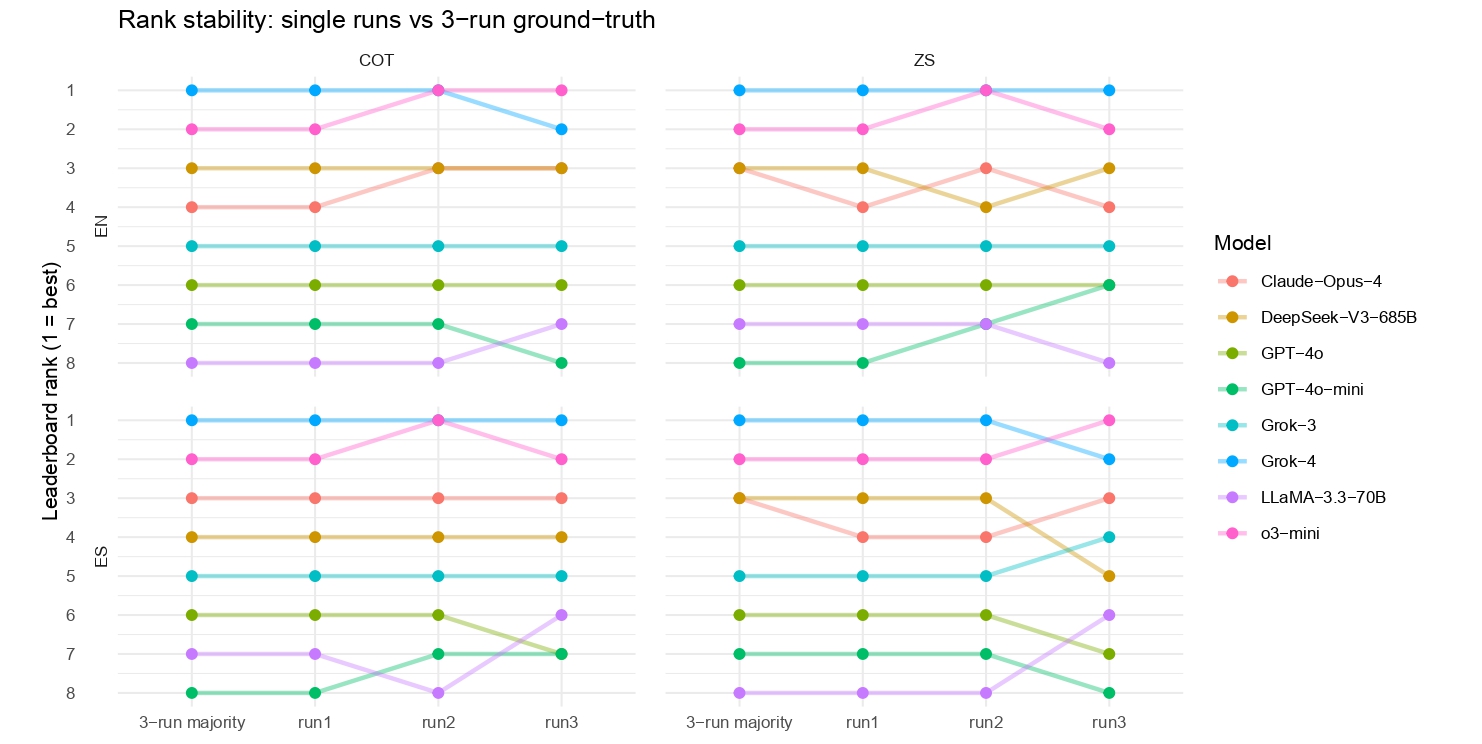} 
    \caption{Rank Stability from Single Runs vs Three-Run Ground Truth. The Y-axis shows leaderboard rank (1 = best) and the X-axis lists the 3-run majority (ground truth) alongside three individual runs. Panels are split by prompt regime (COT left, ZS right) and language (EN top, ES bottom). Colored lines (models as in the legend) connect each model’s rank across runs; horizontal segments indicate agreement with the 3-run majority, while sloped segments mark a rank inversion. }
    \label{fig:rank_stability}
\end{figure}

\subsection*{Sign‑Flip Rate in Pairwise Significance}

Pairwise significance was robust: across the eight model-vs-baseline contrasts, no decision flipped when we moved from a single run to the three-run aggregate (flip-rate = 0\%). This stability is expected given both the large effect sizes (odds ratios $\approx 4$--$40$) and the slice-level reliability: the mean ICC is 0.68, the median 0.70, and the upper half of slices average 0.79. With only $\sim$30\% of the variance attributable to run-to-run noise, even a single run already produces confidence intervals that remain well separated from zero.

Despite the rank inversions described in the previous section, the statistical significance of each model’s advantage over Grok-4 proved robust: across seven pairwise contrasts and three seeds, no confidence interval crossed zero (flip-rate = 0\%). Large effect sizes (odds ratios 6×–40×; see Supplement Table 4: Pairwise Contrast vs Grok-4) keep the significance verdict unchanged even when accuracy oscillates by ±2–3 pp.

\subsection*{Stability Improvement From Additional Runs}

\begin{figure}[H] 
    \centering
    \includegraphics[width=0.7\textwidth]{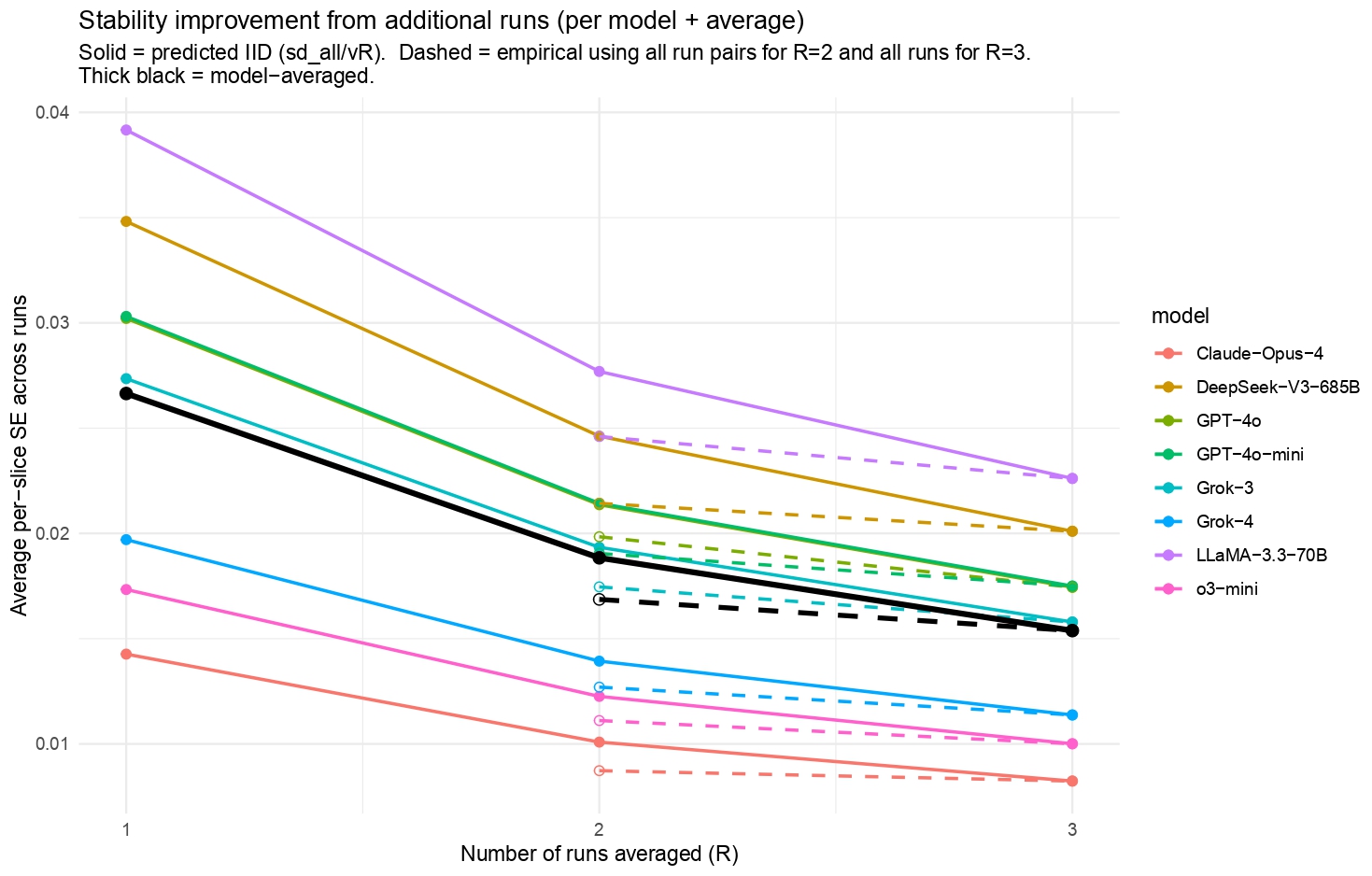} 
    \caption{Stability improvement from additional runs (per model and overall). Solid colored lines show the predicted reduction in the average per-slice standard error (SE) when averaging $R$ stochastic runs, computed as the across-run SD per slice divided by $\sqrt{R}$ (IID assumption). Dashed lines show the empirical SE computed on a fixed set of slices: for $R=2$ we average the SE over all run pairs, and for $R=3$ we use all three runs. The thick black line is the model-average.}
    \label{fig:run_stability}
\end{figure}

Averaging multiple stochastic decodes reduces the standard error (SE) of slice-level accuracy roughly with a $1/\sqrt{R}$ law. Figure 9 overlays per-model trajectories (colored) with the model-averaged trend (thick black).  The predicted IID curves (solid) closely track the empirical estimates (dashed), with only minor, expected deviations due to finite-sample variability.

On average, SE decreases from $0.0288$ at $R=1$ → $0.0285$ at $R=2$ → $0.0273$ at $R=3$, a $-5\%$ total reduction, consistent with diminishing returns. The main practical benefit of repetition is not much narrower confidence intervals (CIs) but more stable rankings—two runs remove $\sim$83\% of single-run inversions (see Fig. 8). We therefore adopt two runs as a cost-effective default; a third run is optional for high-precision reporting.

\section*{Conclusion}

Our work highlights both the progress and the limitations of state-of-the-art reasoning-focused LLMs under rigorous evaluation. Leaderboards often underreport statistical variation, yet our results show that replication is essential for drawing reliable conclusions. We propose practical strategies to determine how many repetitions are sufficient for a given benchmark, balancing robustness with cost-effectiveness.

Across all evaluation settings, Grok-4 achieved the highest aggregate accuracy, closely followed by o3-mini and Claude Opus 4. Grok-4 performance on the AI4Math benchmark was statistically indistinguishable from o3-mini, indicating that superiority is not guaranteed even among flagship systems. Domain-level analyses further revealed complementary strengths—procedural persistence in Grok-4 and spatial visualization in o3-mini—alongside shared weaknesses in area-related and low-information optimization tasks. These findings underscore that no single model dominates across all problem types, and model selection may need to be task-dependent.

Our analysis also shows that newer models exhibit greater self-consistency than earlier generations. However, rank instability remains a serious concern: 83\% of single runs led to rank inversions compared to the three-run aggregate. This demonstrates that single-run leaderboards are unreliable; repetitions are necessary to stabilize rankings, even if confidence intervals remain largely unchanged. Rather than prescribing a fixed number of runs, we emphasize that researchers should establish an appropriate replication threshold tailored to their dataset and evaluation design. Doing so ensures stable rankings without incurring unnecessary computational costs.

Finally, we view this work as part of the democratization of independent LLM evaluation. Diverse, transparent, and open benchmarks are critical for building assessments that are both representative and reproducible. Conversely, limited independent participation risks misaligned incentives and overfitting to narrow metrics. By providing accessible statistical tools and cost-aware replication strategies, we aim to support smaller research teams and strengthen evaluation practices that are rigorous yet widely adoptable. In turn, this contributes to a more reliable and socially grounded foundation for the responsible deployment of AI across scientific, educational, and societal domains.

\section*{Acknowledgments}
We thank Pablo Villalobos for his thoughtful feedback and review of this work, and Jaime Esteban Montenegro Barón for insightful comments on the statistical analysis. We also acknowledge Alison Diaz Cuevas for her contribution during the hackathon at which the original dataset was generated; we regret the earlier omission and are pleased to recognize her contribution here.

\section*{Additional Resources}
Supplementary Figures and Tables can be found at:

\href{https://github.com/malvarado-tech/AI4MATH_2_R_project/tree/main/Supplementary}{$https://github.com/malvarado-tech/AI4MATH\_2\_R\_project/tree/main/Supplementary$}


\end{document}